%% file: sigconf.tex
\documentclass[sigconf]{acmart}

\acmYear{2026}
\acmConference[DAI '26]
  {The Eighth International Conference on Distributed Artificial Intelligence}
  {November 29--December 2, 2026}
  {Hong Kong, China}
\acmBooktitle{The Eighth International Conference on Distributed Artificial
  Intelligence (DAI '26), November 29--December 2, 2026, Hong Kong, China}

\setcopyright{none}
\copyrightyear{}
\acmDOI{}
\acmISBN{}
\usepackage{multirow}
\usepackage{booktabs}
\usepackage{hhline}
\usepackage{latexsym}
\usepackage{amsmath}
\usepackage{enumitem}
\usepackage{pgfplots}
\usepackage{pgfplotstable}
\usepackage{tikz}
\usetikzlibrary{patterns}
\usepackage{graphicx}
\usepackage{algorithm}
\usepackage{algpseudocode}
\usepackage[most]{tcolorbox}
\tcbuselibrary{breakable}
\usepackage{colortbl}
\usepackage{changepage}

\definecolor{lightblue}{RGB}{173, 216, 245}
\definecolor{verylightblue}{RGB}{224, 242, 255}

\newcommand{\best}[1]{\cellcolor{lightblue} #1}
\newcommand{\second}[1]{\cellcolor{verylightblue} #1}

\definecolor{MyDeepGreen}{RGB}{0,180,0}
\definecolor{MyDeepRed}{RGB}{250,0,0}

\newcommand{\ours}{\textsc{AutoMind}}

\newcommand{\tree}{$\mathcal{T}$}
\newcommand{\node}{$\mathcal{N}$}
\newcommand{\parent}{$N_\text{parent}$}
\newcommand{\child}{$N_\text{child}$}
\newcommand{\edge}{$\mathcal{E}$}

\newcommand{\plan}{$p$}
\newcommand{\code}{$\sigma$}
\newcommand{\metric}{$\eta$}

\newcommand{\policy}{$\pi$}
\newcommand{\action}{$\mathcal{A}$}
\newcommand{\adraft}{$\mathcal{A}_\text{draft}$}
\newcommand{\adebug}{$\mathcal{A}_\text{debug}$}
\newcommand{\aimprove}{$\mathcal{A}_\text{improve}$}

\begin{document}

\title{\ours: Adaptive Knowledgeable Agent for Automated Data Science}

\author{Yujie Luo}
\authornotemark[1]
\affiliation{%
  \institution{Zhejiang University}
  \city{Hangzhou}
  \country{China}}
\email{luo.yj@zju.edu.cn}

\author{Yixin Ou}
\authornote{These authors contributed equally to this work.}
\affiliation{%
  \institution{Zhejiang University}
  \city{Hangzhou}
  \country{China}}
\email{ouyixin@zju.edu.cn}

\author{Jingsheng Zheng}
\authornotemark[1]
\affiliation{%
  \institution{Zhejiang University}
  \city{Hangzhou}
  \country{China}}

\author{Lanning Wei}
\affiliation{%
  \institution{Ant Group}
  \city{Hangzhou}
  \country{China}}

\author{Zhuoyun Yu}
\affiliation{%
  \institution{Zhejiang University}
  \city{Hangzhou}
  \country{China}}

\author{Shuofei Qiao}
\affiliation{%
  \institution{Zhejiang University}
  \city{Hangzhou}
  \country{China}}

\author{Jintian Zhang}
\affiliation{%
  \institution{Zhejiang University}
  \city{Hangzhou}
  \country{China}}

\author{Da Zheng}
\authornote{Corresponding authors.}
\affiliation{%
  \institution{Ant Group}
  \city{Hangzhou}
  \country{China}}
\email{zhengda.zheng@antgroup.com}

\author{Yuren Mao}
\affiliation{%
  \institution{Zhejiang University}
  \city{Hangzhou}
  \country{China}}

\author{Yunjun Gao}
\affiliation{%
  \institution{Zhejiang University}
  \city{Hangzhou}
  \country{China}}

\author{Huajun Chen}
\affiliation{%
  \institution{Zhejiang University}
  \city{Hangzhou}
  \country{China}}

\author{Ningyu Zhang}
\authornotemark[2]
\affiliation{%
  \institution{Zhejiang University}
  \city{Hangzhou}
  \country{China}}
\email{zhangningyu@zju.edu.cn}

\renewcommand{\shortauthors}{Luo et al.}

\begin{abstract}
Large Language Model (LLM) agents have shown great potential in addressing real-world data science problems. LLM-driven data science agents promise to automate the machine learning pipeline, yet their real-world effectiveness remains limited. Existing frameworks depend on rigid, pre-defined workflows and inflexible coding strategies, consequently, they excel only on relatively simple problems and fail to capture the empirical expertise that human practitioners bring to complex tasks. In this work, we introduce {\ours}, an adaptive, knowledgeable LLM-agent framework that overcomes these deficiencies through three key advances: (1) a curated expert knowledge base that grounds the agent in domain expert knowledge, (2) an agentic knowledgeable tree search algorithm that strategically explores possible solutions, and (3) a self-adaptive coding strategy that dynamically tailors code generation to task complexity. Evaluations on two automated data science benchmarks demonstrate that {\ours} delivers superior performance versus state-of-the-art baselines. Additional analyses confirm favorable effectiveness, efficiency, and qualitative solution quality, highlighting {\ours} as an efficient and robust step toward fully automated data science. Our code, data and logs for experiments are open-sourced\footnote{\url{https://github.com/zjunlp/AutoMind}}.
\end{abstract}



\maketitle

\input{sections/1_introduction}
\input{sections/2_preliminaries}
\input{sections/3_method}
\input{sections/4_experiments}
\input{sections/5_analysis}
\input{sections/6_related_work}
\input{sections/7_conclusion}


\begin{acks}
We would like to express our sincere gratitude to the anonymous reviewers for their thoughtful and constructive feedback. 
This work was supported by the National Natural Science Foundation of China (No.62576307), the Yongjiang Talent Introduction Programme (2021A-156-G), and the Information Technology Center and State Key Lab of CAD\&CG, Zhejiang University.  This work was supported by Ant Group and Zhejiang University - Ant Group Joint Laboratory of Knowledge Graph.
\end{acks}

\bibliographystyle{ACM-Reference-Format}

\bibliography{custom}

\appendix
\input{appendix/benchmarks}
\input{appendix/environment}
\input{appendix/hyperparameters}
\input{appendix/prompts}

\end{document}

%% file: sections/1_introduction.tex
\section{Introduction}

Data science agents aim to leverage LLM agents to automate data-centric machine learning tasks that begin with task comprehension, data exploration and analysis, advance through feature engineering, and culminate in model selection, training, and evaluation \cite{ds_agent_survey,zhengda_complex_problem_survey,liu2025ml}, serving as a critical component for future AI agents to achieve autonomous scientific discovery.
Many data science-related benchmarks \cite{MLAgentBench,ds-bench,mle-bench,scienceagentbench} have been introduced to provide structured tasks based on real-world challenges, enabling comprehensive evaluation of performance across the full problem-solving pipeline.
Because of the great complexity of these tasks, most existing data science agent frameworks rely on pre-defined workflows and optimize on top of the specific workflow through search and refinement \cite{aide,data-interpreter,automl-agent} or extend to a multi-agent framework to better stimulate the performance of each workflow node \cite{autokaggle}.

However, current data science agents all overlook the fundamental limitations in model capabilities: despite being trained on a massive code-based corpus, the agents inherently lack the rich empirical expertise accumulated by human practitioners in data science tasks \cite{zhengda_complex_problem_survey}.
Moreover, existing data science agents largely employ an inflexible coding strategy and tend to implement code only for relatively simple tasks in practice \cite{ds-agent,autokaggle}.
Yet the complexity of real-world problems requires a dynamic, context-aware coding strategy.
Indeed, addressing truly complex or even cutting-edge tasks that require high levels of creativity and innovation poses significant challenges for data science agents in generating high-quality code appropriately tailored to such complex tasks.

\begin{figure*}[htp]
    \centering
    \includegraphics[width=\linewidth]{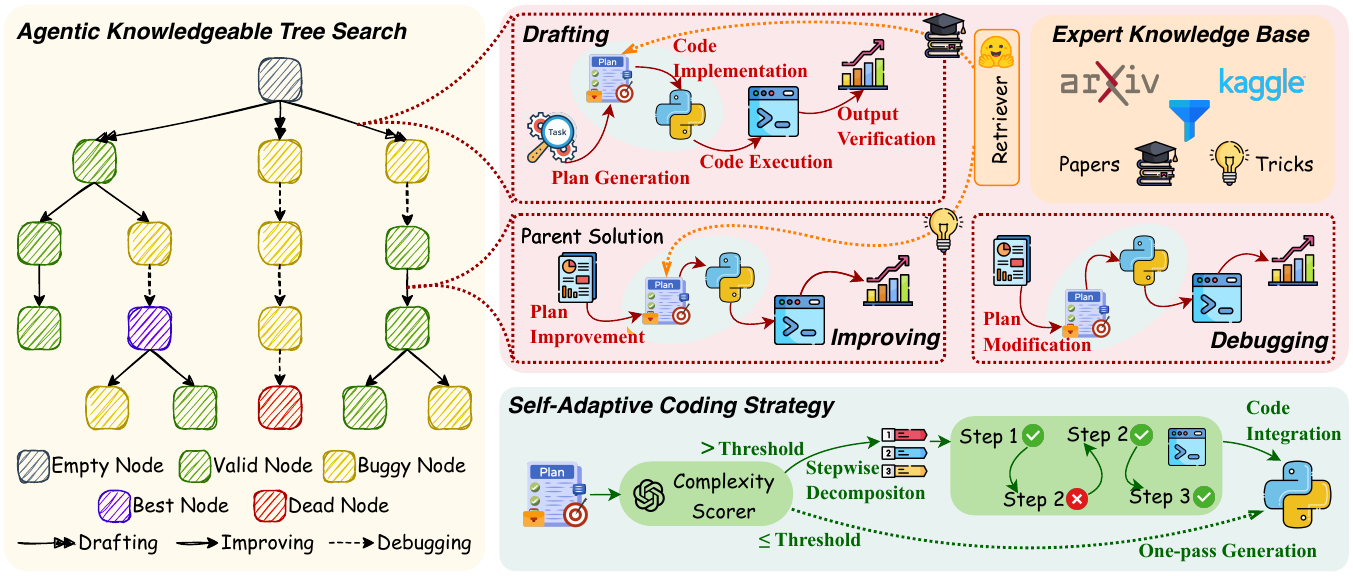}
    \caption{\textbf{The framework of our {\ours}.}The \textbf{Expert Knowledge Base} (upper right) supplies domain papers and competition tricks through a retriever. The \textbf{Agentic Knowledgeable Tree Search} (left) explores the solution space by iteratively expanding a tree via three actions: Drafting creates initial solutions from retrieved papers, Improving refines valid nodes using retrieved tricks, and Debugging fixes buggy nodes through plan modification. The \textbf{Self-Adaptive Coding Strategy} (bottom right) routes each plan through a complexity scorer, where high-complexity plans are decomposed into substeps with per-step verification, and low-complexity plans are generated in a single pass.}
    \label{fig:framework}
\end{figure*}

To tackle these issues, we propose \textbf{{\ours}}, an adaptive, knowledgeable LLM agent framework designed for automated data science challenges.
As illustrated in Figure \ref{fig:framework}, {\ours} introduces three major innovations:
\begin{itemize}[
  leftmargin=*
]
    \item \textbf{Expert Knowledge Base.} An expert-curated repository of data-science knowledge that grounds the agent in empirical best practices, overcoming LLMs' inherent lack of additional human practitioner experience.
    \item \textbf{Agentic Knowledgeable Tree Search Algorithm.} A tree-search strategy that fully leverages the expert knowledge base, enabling the agent to dynamically explore multiple solution pathways and thereby enhance its performance on complex problem-solving tasks.
    \item \textbf{Self-Adaptive Coding Strategy}. A dynamic code-generation mechanism that scales with task complexity, replacing rigid workflows with context-aware implementations and thereby providing flexible, efficient coding solutions across varying levels of implementation difficulty.
\end{itemize}

We evaluate {\ours} on two automated data science benchmarks with two different families of foundation models.
Experimental results show that {\ours} achieves superior performance on both of the benchmarks compared to baselines.
Specifically, on the official MLE-Bench leaderboard, {\ours} surpasses \textbf{40\%} of human participants on average, representing an improvement of \textbf{8.0\%} over the prior state-of-the-art (SOTA).
Moreover, we conduct an in-depth analysis to evaluate the effectiveness and efficiency of {\ours}, and find that {\ours} achieves higher efficiency and lower token costs compared to prior SOTA.

%% file: sections/2_preliminaries.tex
\section{Preliminaries}
\label{sec:preliminaries}

Building on recent successes in integrating tree search strategies with workflows of LLM agents \cite{aide,sela,ai_scientist_v2}, we model LLM agent-driven automated data science as an optimization problem, and apply a tree search algorithm to solve it.

Formally, a possible soluton $s$ for a data science task is defined as a tuple $s=(p,\sigma,\eta)$, where $p$ denotes a textual plan of the proposed approach, $\sigma$ is the Python code snippet, and $\eta$ is the validation metric used to assess the execution results.
Let $\mathcal{S}$ be the space of all possible solutions, and the objective is to find the optimal solution:
\begin{equation}
\begin{split}
    s^*&=\underset{s \in \mathcal{S}}{\arg \max }\,\eta,\\\text { where } \eta^*&=\max \{\eta \mid(p, \sigma, \eta) \in \mathcal{S}\}.
\end{split}
\label{eq:optimal}
\end{equation}

Unlike general agents \cite{react,metagpt,reflexion,fireact}, which conceptualize task solving as a long-horizon decision process aimed at maximizing cumulative reward through action choices based on prior observations, our modeling approach significantly simplifies the objective by directly evaluating and comparing possible solutions for data science tasks.

%% file: sections/3_method.tex
\section{\ours}

\input{algorithms/search_policy}

In this section, we introduce our {\ours}, an adaptive knowledgeable LLM agent framework designed for automated data science challenges.
As illustrated in Figure \ref{fig:framework}, {\ours} introduces three major innovations: an expert knowledge base for data science (\S\ref{sec:knowledge_base}), an agentic knowledge tree search algorithm (\S\ref{sec:tree_search}), and a self-adaptive coding strategy (\S\ref{sec:adaptive_coding}).
First, the agent’s specialized knowledge retriever extracts multiple relevant papers and tricks from the expert knowledge base.
Next, the agentic knowledgeable tree search module initiates an iterative loop in which it selects a parent node according to the search policy, executes an action that synthesizes task information with retrieved knowledge to produce a new solution, and integrates the resulting node into the solution tree.
Concurrently, the self-adaptive coding strategy is invoked during the code implementation stage of each action, reconciling solution complexity with the inherent coding capabilities of LLMs.
Once the iteration limit is reached or the time budget is exhausted, the best node in the solution tree identified by Equation~\eqref{eq:optimal} is selected, submitted, and evaluated as the final solution.
The following sections delve into key implementation details of each component.

\subsection{Expert Knowledge for Data Science}
\label{sec:knowledge_base}
The data science competitions are challenging due to the requirements of high-quality experience in designing effective solutions \cite{mle-bench,automl-agent}.
Using LLMs alone to solve these competitions is challenging due to their reliance on static, pre-trained knowledge, which may lack domain-specific or up-to-date insights.
To address this challenge, we construct a knowledge base based on domain-specific resources, including papers from top-tier conferences and journals, as well as expert-curated insights from top-ranked solutions.

\subsubsection{Knowledge Base Construction}

Following the human expertise, we collect the top-ranked solution tricks from different competitions which expect to fully unleash the power of the existing method, as well as the papers to provide more expressive strategies when solving data science tasks.

In data science tasks, minor yet effective tricks can significantly enhance model performance.
To incorporate such human insights in our framework, we identify all Kaggle competitions with publicly shared solutions\footnote{We collect all Kaggle competitions with solutions using the list from \url{https://github.com/faridrashidi/kaggle-solutions}.}, and then archive both competition descriptions and the content of associated technical forum posts. 
After filtering out invalid competitions and posts, we retain 3,237 public forum posts that offer valid solutions for 455 Kaggle competitions.
Then, different labels describing the competition-oriented technical categories are attached to each solution post for subsequent retrieval.

Besides, the papers accepted after peer-review are high-quality prior knowledge in solving different data science tasks. 
Considering the expressive methods of accepted papers in improving the expressiveness.
To utilize such knowledge, we first collect papers published in top-tier conferences in the recent three years from arXiv, like KDD, ICLR, NeurIPS, ICML, EMNLP, and domain-specific journal like Bioinformatics.
For each paper, the meta information (including title, author, abstract, and keywords) and main content are preserved, from which we obtain the prepared paper knowledge.

\subsubsection{Knowledge Retrieval}
Due to the strong dependence of machine learning tasks on technical approaches, traditional retrieval methods based solely on task-description embeddings are often ineffective.
Since the aforementioned content contains excessive fine-grained details, it is difficult to retrieve precise knowledge for a given task.
To this end, we introduce a hierarchical labeling system to assist knowledge retrieval, filtering, and re-ranking.

For the collection of tricks, we first construct a hierarchy label set based on all collected data science tasks from Kaggle with the assistance of LLMs, and it contains 11 top-level categories and corresponding subcategories (e.g., category Computer Vision and subcategory Image Classification).
Then, to label each trick, {\ours} first selects the most relevant top-level categories and then identifies the most appropriate labels from the corresponding subcategories. 
We ensure labeling quality through a self-consistency strategy with five independent iterations and voting-based arbitration, on which we obtain precise labels for each trick.
Compared with tricks, papers are much more diverse in data and techniques, bringing difficulties in designing hierarchical labels.
Then, we use LLMs to generate a brief summary for each paper from the perspective of data (including type, domain, and dataset name), data science tasks, the proposed techniques and key contributions. 
In this way, papers can be retrieved to solve the different competitions from different perspectives.
In the retrieval stage, the input task is analyzed with the same labeling tricks.
For each label, {\ours} performs a similarity search in the knowledge base to retrieve associated knowledge.
Then, after filtering out solutions or tricks of the same target task to avoid plagiarism, the retrieved results are re-ranked based on the aforementioned label priority order, on which we obtain the final retrieved knowledge.

\subsection{Agentic Knowledgeable Tree Search}
\label{sec:tree_search}

\newcommand{\solution}{$s$}

To facilitate the exploration of possible solutions, we model the search space as a solution tree {$\mathcal{T}=(\mathcal{N}, \mathcal{E})$}.
Each node {$N\in\mathcal{N}\subset\mathcal{T}$} corresponds to a unique solution $s=(p,\sigma,\eta)$ as formalized in \S\ref{sec:preliminaries}, while each eage {$E=(N_\text{parent},N_\text{child})\in\mathcal{E}\subset\mathcal{T}$} corresponds to the specific action applied to the parent node $N_\text{parent}$ that produces the child node $N_\text{child}$.
As illustrated in Figure \ref{fig:framework}, the search tree is initialized as a single empty node $N_{\text{empty}}$, after which {\ours} begins iteratively exploring the solution space.
At the start of each iteration, the search policy {\policy} receives the current state of the {\tree}, selects one node as the {\parent} according to the search policy, and specifies an action that generates a {\child} and launches the next iteration of the exploration.
Next, we examine the core components of the exploration framework in greater detail.

\paragraph{Solution Nodes (\node)}

Each solution node $N\in\mathcal{N}$ consists of following information:
\begin{itemize}[
  leftmargin=*
]
    \item \textbf{Plan {\plan}}: an end-to-end textual solution plan typically comprises sequential stages including data pre-processing, feature engineering, model training, and prediction.
    \item \textbf{Code {\code}}: a Python implementation of the solution plan \plan.
    \item \textbf{Output $o$}: the terminal output generated during the execution of code {\code}, which serves as a diagnostic feedback signal.
    \item \textbf{Metric {\metric}}: the task-specific validation score extracted from the terminal output $o$.
\end{itemize}
As shown in Figure~\ref{fig:framework}, solution nodes are classified as either \textit{valid} nodes $\mathcal{N}_{\text {valid}}$ or \textit{buggy} nodes $\mathcal{N}_{\text {buggy}}$, based on whether their metrics can be correctly computed.
Additionally, if a buggy node reaches the pre-defined max debug depth, it will be marked as a \textit{dead} node, which will not be further selected by the search policy.
The \textit{best} node $N_{\text{best}}$ is marked as the valid node with the optimal validation metric in the {\tree}.

\paragraph{Action Edges (\edge)}

The action space of {\ours} consists of three distinct operations: \textbf{Drafting} \adraft, \textbf{Improving} \aimprove, and \textbf{Debugging} \adebug.
The action specified at each iteration is based on the type of the parent node {\parent} selected by the search policy \policy, which can be {\adraft}, {\aimprove} or {\adebug} if the {\parent} is empty, valid, or buggy respectively.
As illustrated in Figure \ref{fig:framework}, each action goes through a similar pipeline processed through plan generation, code implementation, code execution, and output verification.
However, different types of actions vary primarily in the specific inputs provided for the plan generation stage.
In the {\adraft}, {\ours} synthesizes the task description with relevant papers retrieved from the expert knowledge base to formulate an initial solution.
In the {\aimprove}, {\ours} is provided with the valid {\parent}—consisting of plan {\plan}, code {\code} and output $o$—as well as tricks retrieved from the expert knowledge base, and is instructed to improve the plan accordingly.
In the {\adebug}, {\ours} receives only the buggy {\parent} and is instructed to modify the plan to resolve the bug.
Procedures for code implementation, execution, and output verification are uniformly applied across all action types.
Once an action completes, the resulting solution {\solution} is encapsulated as a new node {\child} within the {\tree} as the child of the {\parent} selected by the search policy {\policy}.

\paragraph{Search Policy (\policy)}

The algorithm for search policy is driven by simple yet effective heuristic rules with probabilistic hyperparameters.
Specifically, the policy first ensures the maximum number of draft nodes $C_\text{init}$ is met, which splits the search tree into multiple branches, to lay the foundation for further exploration:
\begin{equation*}
\begin{split}
    \pi_0(\mathcal{T})=\begin{cases}\left(N_{\text {empty }}, \mathcal{A}_{\text {draft }}\right) & \text {if } \lvert\mathcal{N}_{\text {draft }}\rvert<C_{\text {init }} \\ \pi_1(\mathcal{T}) & \text {otherwise}\end{cases},
\end{split}
\end{equation*}
where $\mathcal{N}_{\text {draft }}$ is the set of draft nodes in the {\tree}.
It then prioritizes debugging the buggy leaf nodes with a heuristic probability $H_{\text{debug}}$, thereby enabling the exploration of the potential within existing buggy solutions and facilitating more valid comparisons among all solutions in the {\tree}:
\begin{equation*}
\begin{split}
    \pi_1(\mathcal{T})=\begin{cases}\left(N_{\text {buggy}}, \mathcal{A}_{\text {debug }}\right) & \text {if }p_1<H_{\text {debug }} \\&\text {and } \mathcal{N}_{\text {buggy }} \neq \emptyset  \\ \pi_2(\mathcal{T}) & \text {otherwise}\end{cases},\\\text{where } p_1\sim U(0,1) \text{ and } N_{\text {buggy}}\sim U(\mathcal{N}_{\text {buggy }}).
\end{split}
\end{equation*}
Subsequently, there is a heuristic probability of $H_{\text{greedy}}$ that the current best node with the optimal metric in the {\tree}, will be selected for improvement; and the remaining $1-H_{\text{greedy}}$ are allocated to the selection of other valid nodes for improvement, helping to mitigate the risk of overlooking potentially better solutions:
\begin{equation*}
\begin{split}
    \pi_2(\mathcal{T})=\begin{cases}\left(N_{\text {best }}, \mathcal{A}_{\text {improve }}\right) & \text {if } p_2<H_{\text {greedy }} \\&\text {and } \mathcal{N}_{\text {valid }} \neq \emptyset \\ \left(N_{\text {valid}}, \mathcal{A}_{\text {improve }}\right) &  \text {otherwise}\end{cases},\\\text{where } p_2\sim U(0,1) \text{ and } N_{\text {valid}}\sim U(\mathcal{N}_{\text {valid}}).
\end{split}
\end{equation*}
If no further nodes remain to improve or debug, the policy will expand the search branches by creating a new draft node.
We provide the complete pseudocode for the search policy {\policy} in Algorithm~\ref{algo:search_policy}.

\subsection{Self-Adpative Coding Strategy}
\label{sec:adaptive_coding}
\input{tables/main}

To cope with the spectrum of data science workloads from straightforward machine learning models to multi-stage, state-of-the-art architectures, we introduce within {\ours} a self-adaptive coding mechanism, reconciling solution complexity with the coding capabilities of LLMs, as illustrated in Figure \ref{fig:framework}.

During the code implementation stage of an action, {\ours}  follows a list of professional rubrics to score the overall complexity of the solution plan on a five-point scale.
When this score falls below a preset threshold—indicating that the agent regards the plan as straightforward—the agent implements the entire code for the plan in one pass to maximize efficiency.
However, if the score exceeds this threshold, the agent switches to a stepwise strategy by decomposing the plan into sequential substeps and incorporating execution feedback at each substep.
Specifically, for each substep, the agent performs an Abstract Syntax Tree (AST) check and then executes the code in a terminal session.
If the tests pass, the agent advances to the next substep; otherwise, the agent regenerates the substep’s implementation using the error messages as feedback.
This loop repeats until either the tests succeed for all substeps, after which the agent integrates the substeps’ code into a complete implementation; or a predefined retry limit is reached for any substep, forcing the agent to abandon the current plan.

%% file: algorithms/search_policy.tex
\begin{algorithm*}[!htbp]
\caption{Search Policy {\policy} in {\ours}}
\begin{algorithmic}[1]
\Require{\tree} \Comment{Current state of solution tree}
\Require{$C_\text{init}$} \Comment{Hyper-parameter of the counts of initial draft nodes}
\Require{$H_\text{debug}$} \Comment{Hyper-parameter of the probability of whether debug a node}
\Require{$H_\text{greedy}$} \Comment{Hyper-parameter of the probability of whether select the best node}
\Ensure{({\node}, {\action})} \Comment{Select a parent node and specify an action applied on it}
\State $C_\text{draft} \gets \text{Get the counts of draft nodes in \tree}$
\If{$C_\text{draft} < C_\text{init}$} 
    \Return ($\mathcal{N}_\text{empty}$, $\mathcal{A}_\text{draft}$) \Comment{Draft a new solution}
\EndIf

\State $p_\text{debug} \gets \text{Get a random floating-point number from 0 to 1}$
\State $\mathcal{N}_\text{buggy} \gets \text{Get a random buggy leaf node in \tree}$
\If{$p_\text{debug} < H_\text{debug}$ \textbf{and} $\mathcal{N}_\text{buggy}$ is not None} 
    \Return ($\mathcal{N}_\text{buggy}$, $\mathcal{A}_\text{debug}$) \Comment{Debug a buggy solution}
\EndIf

\State $p_\text{greedy} \gets \text{Get a random floating-point number from 0 to 1}$
\State $\mathcal{N}_\text{best} \gets \text{Get the best node in the solution tree}$
\State $\mathcal{N}_\text{valid} \gets \text{Get a random valid node in the solution tree}$
\If{$p_\text{greedy} < H_\text{greedy}$ \textbf{and} $\mathcal{N}_\text{best}$ is not None} 
    \State \Return ($\mathcal{N}_\text{best}$, $\mathcal{A}_\text{improve}$) \Comment{Improve the current best valid solution}
\ElsIf{$p_\text{greedy} \ge H_\text{greedy}$ \textbf{and} $\mathcal{N}_\text{valid}$ is not None} 
    \State \Return ($\mathcal{N}_\text{valid}$, $\mathcal{A}_\text{improve}$) \Comment{Mitigate getting trapped in local optima}
\EndIf

\State \Return ($\mathcal{N}_\text{empty}$, $\mathcal{A}_\text{draft}$) \Comment{No solution to debug or improve, draft a new solution}
\end{algorithmic}
\label{algo:search_policy}
\end{algorithm*}

%% file: tables/main.tex
\begin{table*}[t]
\centering
\caption{\textbf{Main results.} For MLE-Bench, we report both the best@3 and avg@3 win rates against human participants. For the OAG and BELKA competitions, we report the avg@3 task metrics, which are AUC and AP, respectively. The \colorbox{lightblue}{best} and \colorbox{verylightblue}{suboptimal} results for each task are highlighted. $^\ddag$ indicates that the results are borrowed from the grading reports of previous work.}
\label{tab:main}
\resizebox{\textwidth}{!}{
\begin{tabular}{c|c|cc|cc|cc|cc|ccc}
\toprule[1.2pt]
\multirow{3}{*}{\textbf{Method}} & \multirow{3}{*}{\textbf{Backbone}} & \multicolumn{8}{c|}{\textbf{MLE-Bench}} & \multicolumn{3}{c}{\textbf{Top AI Competitions}} \\ \cline{3-10}\cline{11-13}
&  & \multicolumn{2}{c|}{\textit{Easy}} & \multicolumn{2}{c|}{\textit{Medium}} & \multicolumn{2}{c|}{\textit{Hard}} & \multicolumn{2}{c|}{\textit{Overall}} & \multirow{2}{*}{\textit{OAG}} & \multirow{2}{*}{\textit{BELKA}} & \multirow{2}{*}{\textit{Overall}} \\ \cmidrule(lr){3-10}
&  & Best@3 & Avg@3 & Best@3 & Avg@3 & Best@3 & Avg@3 & Best@3 & Avg@3 &  & & \\
\midrule[1pt]
MLAB & \texttt{GPT-4o}$^\ddag$ & 0.22 & 0.13 & 0.08 & 0.04 & 0.03 & 0.02 & 0.11 & 0.06 & - & - & -\\ \midrule[1pt]
OpenHands & \texttt{GPT-4o}$^\ddag$ & 0.48 & 0.28 & 0.08 & 0.07 & 0.17 & 0.08 & 0.24 & 0.15 & - & -  & -\\ \midrule[1pt]
\multirow{3}{*}{AIDE} & \texttt{GPT-4o}$^\ddag$ & 0.71 & 0.53 & 0.26 & 0.15 & 0.13 & 0.07 & 0.37 & 0.25 & - & - & - \\
& \texttt{o3-mini}
& 0.53 & 0.44 & 0.22 & 0.19 & 0.27 & 0.20 & 0.34 & 0.28 & \second 0.56 & 0.09 & 0.33 \\
& \texttt{DeepSeek-V3} & 0.79 & 0.58 & 0.36 & 0.31 & 0.28 & 0.21 & 0.48 &  0.36 & 0.52 & 0.33 & 0.43 \\ \midrule[1pt]
\begin{tabular}[c]{@{}c@{}}\textbf{\ours}\\ w/o Knowledge\end{tabular} & \texttt{DeepSeek-V3} & \best \textbf{0.85} & 0.54 & 0.31 & 0.21 & \second 0.40 & \second 0.24 & 0.52 & 0.33 & 0.50 & 0.19 & 0.35 \\ \midrule[1pt]
\multirow{2}{*}{\textbf{\ours}} & \texttt{o3-mini} & 0.81 & \second 0.60 & \best \textbf{0.50} & \second 0.33 & 0.32 & 0.23 & \second 0.54 & \second 0.39 & 0.55 & \best \textbf{0.44} & \best \textbf{0.50} \\
& \texttt{DeepSeek-V3} & \second 0.83 & \best \textbf{0.65} & \second 0.48 & \best \textbf{0.33} & \best \textbf{0.44} & \best \textbf{0.26} & \best \textbf{0.58} & \best \textbf{0.41} & \best \textbf{0.58} & \second 0.39 & \second 0.49 \\
\bottomrule[1.2pt]
\end{tabular}
}
\end{table*}


%% file: sections/4_experiments.tex
\section{Experiments}

\subsection{Experimental Setup}

\paragraph{Backbone Models}

In the main experiment, we evaluate the agents by systematically varying the backbone models. Specifically, we test three representative models: \texttt{GPT-4o}\footnote{gpt-4o-2024-08-06} and \texttt{o3-mini}\footnote{o3-mini-2025-01-31} from OpenAI, and \texttt{DeepSeek-V3}\footnote{DeepSeek-V3-0324} from DeepSeek.

\paragraph{Baseline Agents}

Given the substantial computational resources necessary for baseline reproduction, the results from MLAB~\cite{MLAgentBench}, OpenHands~\cite{OpenHands}, and AIDE~\cite{aide} on MLE-Bench utilizing \texttt{GPT-4o} as the backbone are adopted from prior work\footnote{\url{https://github.com/openai/mle-bench/tree/main/runs}} to serve as initial baselines.
Subsequently, to facilitate a broader quantitative comparison, we re-execute AIDE using \texttt{o3-mini} and \texttt{DeepSeek-V3} as alternative backbones within the scope of the main experiment.
For each evaluation task, the agents are allocated a 24-hour time budget to produce their final submission.
We repeat all experiments with 3 runs per task.
Given the inherent high variance of agents when performing these long-duration tasks, we report the \textbf{best@3} and \textbf{avg@3} performance metrics to provide a more robust and reliable assessment.
We provide detailed runtime environment settings in Appendix~\ref{app:environment} and necessary hyperparameters for reproduction in Appendix~\ref{app:hyperparameters}.

\paragraph{Runtime Environment Settings}

In our experiments, LLM agents are loaded into an Ubuntu 20.04 Docker container containing the dataset prepared for each task, and an Anaconda environment pre-installed with standard Python packages for machine learning (e.g., PyTorch, scikit-learn), thereby providing all requisite dependencies for code implementation and execution.
The container runs on a compute node with 48 vCPUs, 448GB RAM, 9.6TB SSD storage, and a single NVIDIA GeForce RTX 3090 GPU, all of which are fully accessible to the agents.
For each of the task across MLE-Bench and Top AI Competitions, LLM agents are allotted a 24-hour wall-clock budget to make final submission.
We repeat all experiments with 3 runs per task to report the best@3 and avg@3 performance metrics.

\subsection{Benchmarks}

\paragraph{MLE-Bench}

We select MLE-Bench~\cite{mle-bench}, which consists of 75 offline Kaggle competitions for evaluating LLM agents, as part of our test benchmarks.
We further apply a rule-based filtering to the tasks in MLE-Bench as detailed in Appendix~\ref{app:benchmarks}.

Consequently, we obtain a lite version of MLE-Bench conssiting of 15 tasks, which are split into \textit{Easy}, \textit{Medium} and \textit{Hard} tiers based on human experience and the results of previous works~\cite{mle-bench}.
We include a full list of tasks in MLE-Bench used for evaluation in Appendix~\ref{app:benchmarks}.
We assess agent performance by comparing their submissions with official competition leaderboards and report the win rate, which is defined as the proportion of human participants whose scores are surpassed by agents.

\paragraph{Top AI Competions}

An inspection of the original MLE-Bench reveals that most tasks were curated before 2023, with several classical machine learning tasks dating to 2018 or earlier.
Considering the fact that foundation models are likely seen corresponding tasks during pre-training, we supplement our evaluation with two tasks drawn from recent top AI competitions.
Specifically, we include the WhoIsWho-IND track of the Open Academic Graph (OAG) Challenge at KDD Cup 2024 \cite{oag}, evaluated by the area under the ROC curve (AUC), and the BELKA Challenge at the NeurIPS 2024 Competition \cite{belka}, assessed by average precision (AP).
We employ the official task metrics, evaluating the experimented agents directly by their raw scores.
More details about the competitions are shown in Appendix \ref{app:benchmarks}.

\subsection{Main Results}

\begin{figure*}[htp]
    \centering
    \includegraphics[width=\linewidth]{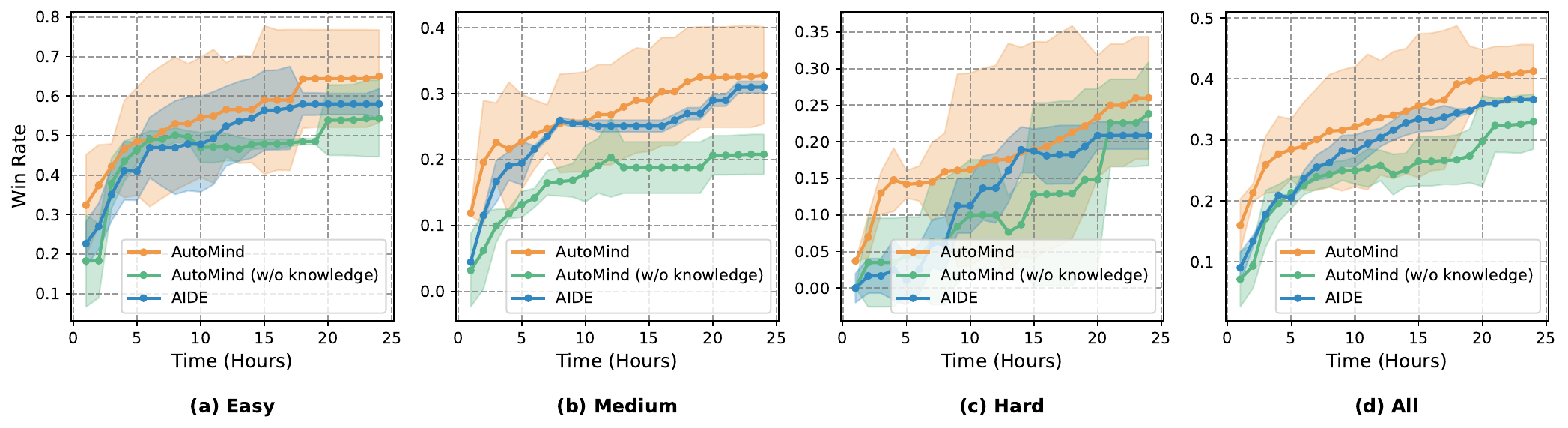}
    \caption{\textbf{Test time scaling results on MLE-Bench.} We record hourly snapshots of the percentage of human participants surpassed by the agent's best solution over a 24-hour time budget in experiments with \texttt{DeepSeek-V3}.}
    \label{fig:time}
\end{figure*}

\begin{figure}[htp]
  \centering
  \includegraphics[width=0.9\linewidth]{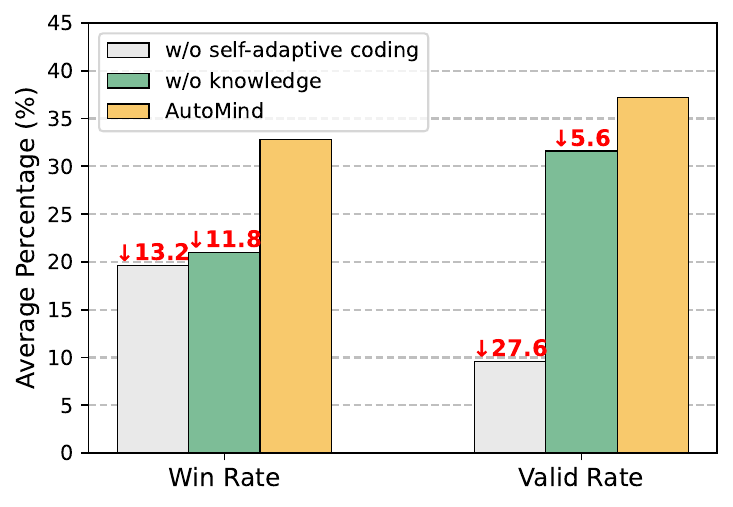}
  \caption{\textbf{Abaltion studies} on \texttt{DeepSeek-V3} for {\ours} on the \textit{Medium} split of MLE-Bench. \textbf{Win Rate} represents the percentage of human participants surpassed by the agent on the official leaderboard. \textbf{Valid Rate} represents the percentage of valid submissions among all solutions the agent makes within a 24-hour time budget.}
  \label{fig:ablation}
\end{figure}

As shown in Table~\ref{tab:main}, {\ours} consistently exceeds all baseline methods under both the best@3 and avg@3 settings.
We find that {\ours} (\texttt{o3-mini}) and {\ours} (\texttt{DeepSeek-V3}) outperform 38.7\% and 41.2\% of human participants on the official leaderboard of MLE-Bench respectively, representing performance gains of 11.0\% and 5.2\% over the prior SOTA AIDE under the avg@3 setting.
Moreover, {\ours} exhibits remarkable superiority under the best@3 setting, achieving win rate improvements of 20.2\% with \texttt{o3-mini} and 10.6\% with \texttt{DeepSeek-V3} over the prior SOTA.
Moreover, {\ours} exhibits remarkable superiority on the \textit{Hard} split of MLE-Bench, achieving Beats (\%) improvements of 25.9\% with \texttt{o3-mini} and 20.3\% with \texttt{deepseek-v3} over the prior SOTA.
Notably, even running {\ours} without the expert knowledge base still outperforms the prior SOTA by 8.8\% on the Beats (\%) metric.
Across both the OAG and BELKA challenges in Top AI Competitions, {\ours} delivers performance that is at least on par with, and in most cases exceeds prior SOTA.
Particularly, {\ours} (\texttt{o3-mini}) achieves an average precision of 0.44 on the BELKA challenge, representing a 0.35 absolute improvement over the prior SOTA.

%% file: sections/5_analysis.tex
\section{Analysis}
\subsection{Ablation Study}

\begin{table}[!t]
    \centering
    \caption{\textbf{Knowledge Retrieval Strategy Comparison.} Evaluation on 15 data science tasks, top-5 solutions retrieved per task (75 total retrievals).}
    \label{tab:retrieval}
    \resizebox{\columnwidth}{!}{
    \begin{tabular}{lcccc}
        \toprule
        \textbf{Method} & \textbf{Precision} & \textbf{Recall} & \textbf{F1} & \textbf{Relevant} \\
        \midrule
        Embedding-only & 52\% & 68\% & 0.59 & 51/75 \\
        Label-only & 78\% & 42\% & 0.55 & 32/75 \\
        \textbf{Ours (Hybrid)} & \textbf{84\%} & \textbf{74\%} & \textbf{0.79} & \textbf{56/75} \\
        \bottomrule
    \end{tabular}
    }
\end{table}

To validate the effectiveness of our design, we conduct ablation experiments for {\ours} (\texttt{DeepSeek-V3}) on the \textit{Medium} split of MLE-Bench, separately disabling two principal components in {\ours}: expert knowledge base and self-adaptive coding strategy.

\paragraph{Expert knowledge provides additional effective supervision for agentic tree search.} 

When {\ours} is run without access to the expert knowledge base, the agent is forced to rely exclusively on its internal knowledge to draft and improve the solutions.
The results in Figure \ref{fig:ablation} demonstrate that ablating the expert knowledge base leads to respective declines of 11.8\% and 5.6\% in the win rate and valid rate metrics.
Figure \ref{fig:time} presents hourly snapshots of the win rate metric over a 24-hour time budget, demonstrating that {\ours} equipped with the expert knowledge base consistently outperforms the variant without it.
We attribute these performance gains to the integration of expert knowledge, which imposes additional constraints on the agent’s solution search space and acts like a "shortcut" to the collective craft of experienced Kaggle grand-masters and recent data science literature.
By leveraging human-validated knowledge, {\ours} reduces its reliance on limited internal knowledge of backbone LLMs, avoids rediscovering effective ideas from scratch in the limitd time budget, and focuses exploration within a more promising solution space.

\paragraph{Self-adaptive coding provides robust support for the implementation of more complex plans.}

We ablate the self-adaptive coding mechanism by completely replacing it with a one-pass coding strategy during the code implementation stage of {\ours}.
The results in Figure \ref{fig:ablation} demonstrate that, replacing the self-adaptive coding mechanism with a one-pass strategy leads to respective declines of 13.2\% and 27.6\% in the win rate and valid rate metrics, highlighting its significant limitations in addressing complex tasks and plans.
We attribute this decline to the limited coding capacity of the backbone LLMs, which proves insufficient to tackle complex tasks and plans in one-pass generation.
By applying stepwise decomposition of complex plans and integrating AST check with execution feedback, error accumulation in the early segments of code generated by the one-pass strategy can be minimized, thereby preserving the efficient execution of subsequent code segments.
As for simpler tasks and plans, the self-adaptive coding strategy inherently permits the utilization of one-pass generation, thereby striking a balance between efficiency and robustness in {\ours}.

\subsection{Effectiveness Analysis}

\paragraph{Knowledge Retrieval Effectiveness}

Our knowledge retrieval mechanism combines manually-defined categories, LLM-driven fine-grained labeling, and embedding-based re-ranking to balance rigor and coverage. To validate this hybrid approach, we conduct a comparative evaluation on 15 data science tasks, retrieving top-5 solutions per task (75 total retrievals). As shown in Table \ref{tab:retrieval}, our hybrid approach significantly outperforms both embedding-only and label-only baselines across all metrics. The embedding-only baseline (MiniLM) achieves high recall (68\%) but low precision (52\%), retrieving many irrelevant solutions. Label-only matching achieves higher precision (78\%) but retrieves fewer relevant items (Recall 42\%). Our hybrid strategy achieves the best balance: Precision 84\%, Recall 74\%, F1 0.79, with 56 out of 75 relevant solutions retrieved. This result validates our insight that labels constrain the solution space to relevant technique routes while embeddings ensure semantic alignment, instantiating the prior theoretical findings of HyDE \cite{Gao2022PreciseZD} and Query2doc \cite{Wang2023Query2docQE}.

\paragraph{Complexity Scoring and Adaptive Routing}
\begin{figure}[!t]
    \centering
    \includegraphics[width=0.8\columnwidth]{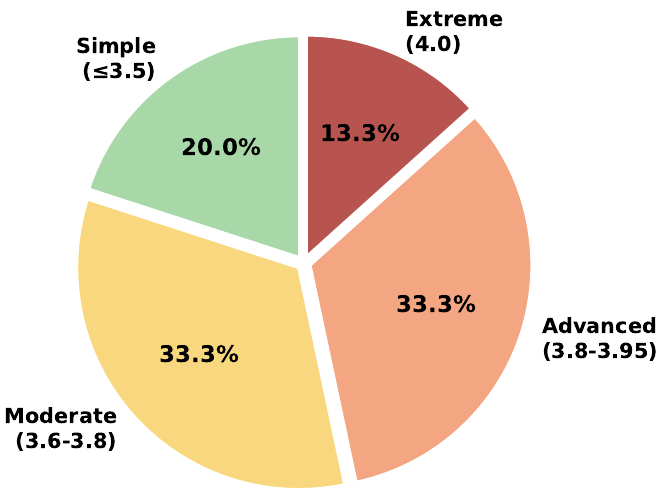}
    \caption{\textbf{Complexity Score Distribution.}}
    \label{fig:complexity}
    \vspace{-6pt}
\end{figure}

To enable self-adaptive coding decisions, we apply rubric-based LLM-as-judge assessment, inspired by the validated paradigm from \cite{Zheng2023JudgingLW}. We score 15 MLE-Bench tasks using a 5-point complexity rubric (averaging 3 runs with \texttt{DeepSeek-V3}). The complexity scores clearly stratify tasks by difficulty. As shown in Figure \ref{fig:complexity}, 2 tasks score 4.0 (lmsys-arena, mlsp-birds), 5 tasks score 3.8-3.95 (aptos2019, volcanic, tgs-salt, google-quest, ventilator), 5 tasks score 3.6-3.8 (ny-taxi, spooky, stanford, statoil, tf-speech), and 3 tasks score $\leq 3.5$ (denoising, us-patent, random-pizza). Our adaptive routing assigns one-pass generation to tasks scoring $\leq 3.5$ (simpler) and stepwise decomposition to tasks scoring $> 3.5$ (complex). This decision boundary is empirically validated by the ablation: forcing one-pass on all tasks reduces Valid Rate by 27.6\% and Win Rate by 13.2\%, directly evidencing that the complexity-driven switch meaningfully identifies and handles tasks of varying difficulty.
\input{tables/tokens}

\subsection{Efficiency Analysis}
\begin{figure*}[!t]
    \centering
    \includegraphics[width=\linewidth]{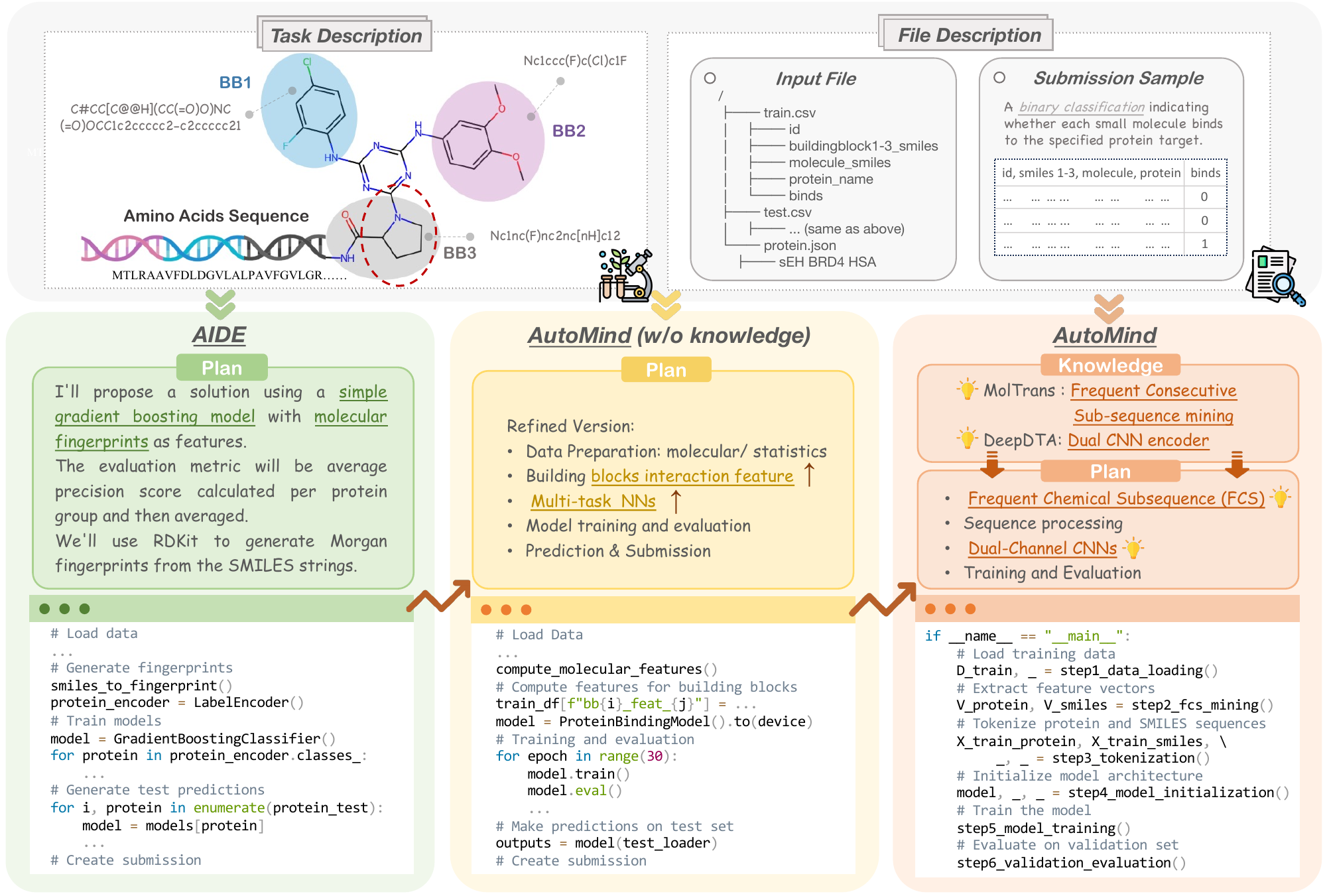}
    \caption{A running case on the BELKA challenge. We compare the proposed solution plans and corresponding code implementations generated by both AIDE and {\ours}.}
    \label{fig:case}
\end{figure*}
\paragraph{Test-Time Scaling}

To assess the efficiency of different agent frameworks, we investigate test-time scaling by tracking the performance of both {\ours} and the prior SOTA AIDE over a 24-hour time budget in experiments with \texttt{DeepSeek-V3}.
As shown in Figure \ref{fig:time}, both agents are able to progressively improve their solutions as the available test-time increases.
Notably, {\ours} achieves the prior SOTA’s 24-hour performance within an average of 15 hours, representing a 60\% improvement in time efficiency.

\paragraph{Token Costs}
We quantifiy the cumulative token costs at the time by which each agent framework achieves the prior SOTA’s 24-hour performance.
As shown in Table \ref{tab:tokens}, owing to the efficiency improvements, {\ours} achieves a 9.6\% reduction in token costs.

\subsection{Case Study}

As shown in Figure \ref{fig:case}, we provide a case study on the BELKA task to verify the effectiveness of \ours.
During execution, {\ours} first retrieves papers MolTrans~\cite{huang2021moltrans} and DeepDTA~\cite{ozturk2018deepdta} from the knowledge base, derives a frequent‑subsequence mining strategy with dual‑CNN blocks inspired by them, and then generates and runs code to implement the plan.
On the contrary, {\ours} (w/o knowledge) focuses on extracting the statistical features of molecules and only adopts the simple MLPs to predict the binding probability.
As for AIDE, the final solution employs a naive gradient boosting model, which is inadequate to tackle such a complex task.
Compared with AIDE and {\ours} (w/o knowledge), {\ours} could retrieve the potential papers and design a more expressive model for complex tasks, the higher performance could demonstrate the effectiveness of the constructed knowledge base and retrieval strategy.

%% file: tables/tokens.tex
\begin{table}[tp]
    \centering
    \caption{\textbf{Token costs across all MLE-Bench tasks.} Input, output, and total token costs for experiments with \texttt{DeepSeek-V3}, reported in millions of tokens.}
    \label{tab:tokens}
    \setlength{\tabcolsep}{1mm}
    \resizebox{0.8\linewidth}{!}{
    \begin{tabular}{lccc}
    \toprule[1.2pt]
    \textbf{Agents} & \textbf{Input} & \textbf{Output} & \textbf{Total} \\
    \midrule[1pt]
    \textbf{AIDE} (24h)  & 2.27 {\small $\pm$ 0.28} & 0.22 {\small $\pm$ 0.03} & 2.49 {\small $\pm$ 0.31} \\
    \midrule
    \textbf{\ours} (15h) & 2.15 {\small $\pm$ 0.24} & 0.27 {\small $\pm$ 0.04} & 2.25 {\small $\pm$ 0.27} \\
    \bottomrule[1.2pt]
    \end{tabular}
    }
\end{table}

%% file: sections/6_related_work.tex
\section{Related Work}
\paragraph{LLM Agents.}
LLMs, with excellent reasoning \cite{shuofei-reasoning-survey,sun2025survey,long-cot-survey} and planning \cite{planning-survey,plangenllms} abilities, are becoming the central control components of AI agents \cite{renda-agent-survey,fudan-agent-survey} and have been increasingly applied in software engineering \cite{metagpt,swe-agent,swe-rl}, deep research \cite{webthinker,deepresearcher}, GUI manipulation \cite{os-atlas,autowebglm,os-agent-survey}, scientific discovery \cite{scienceagentbench,data-interpreter,automl-agent}, embodied intelligence \cite{progprompt,llm-planner}, etc.
Most current LLM agent frameworks rely on two paradigms.
One is the training-free general architecture that depends on the strong capabilities of foundation models and carefully customized workflows~\cite{metagpt,automl-agent,autokaggle}.
The other involves fine-tuning models in specific fields.
Previous studies mainly focused on imitation learning based on a large amount of trajectory data \cite{fireact,agenttuning,os-atlas,autoact}.
However, with the emergence of GRPO-like algorithms \cite{grpo,dapo,vapo}, models can now learn to complete target tasks through self-exploration with rule-based rewards \cite{r1-searcher,search-r1,swe-rl,retool}.

\paragraph{LLM Agents for Data Science.}
Data science agents aim to leverage LLMs to automate data-centric machine learning tasks, including data analysis, data modeling, and data visualization, serving as a critical component for future AI agents to achieve autonomous scientific discovery.
Most existing approaches decompose data science tasks into distinct subtasks with clear boundaries based on human expertise, executing them as workflows within single or multiple agents \cite{data-copilot,ds-agent,autokaggle}.
Furthermore, a series of studies \cite{data-interpreter,aide,automl-agent,sela} employ reflection and search-based optimization.
However, these methods overlook fundamental limitations in model capabilities: despite being trained on massive code datasets, the models inherently lack the rich empirical expertise accumulated by human practitioners in data science.
To integrate human expertise, DS-Agent \cite{ds-agent} adopts a knowledge-based approach by collecting expert Kaggle solutions and applying case-based reasoning to adapt these legacy solutions to new tasks.
Additionally, the inherent complexity of data science task types necessitates diverse problem-solving strategies, whereas current solutions predominantly apply uniform approaches across all tasks.
To address these gaps, this paper proposes to enhance agent capabilities by incorporating human expertise (from research papers, Kaggle competitions, etc.) as an expert knowledge base, while implementing dynamic coding strategy selection mechanisms to adapt to different task requirements.

%% file: sections/7_conclusion.tex
\section{Conclusion}
We introduce {\ours}, an adaptive, knowledge‐driven LLM‐agent framework tailored for automated data science.
By integrating three key innovations, including an expert knowledge base curated for data science, an agentic knowledgeable tree‐search algorithm, and a self‐adaptive coding strategy, {\ours} delivers superior performance versus state-of-the-art baselines on two automated data science benchmarks.
Our experiments validate the effectiveness of {\ours}'s design and quantify its efficiency improvements, highlighting {\ours} as an efficient and robust step toward fully automated data science.


%% file: appendix/benchmarks.tex
\section{Benchmarks}
\label{app:benchmarks}

\input{tables/mlebench}

\subsection{MLE-Bench}
The original MLE-Bench~\cite{mle-bench} consists of 75 offline Kaggle competitions for evaluating LLM agents.
We apply a rule-based filtering to the tasks in MLE-Bench.
Specifically, we first exclude the tasks for which no valid submission can be made from the prior SOTA, thereby eliminating tasks that might be excessively difficult or ill-posed for LLM agents.
From the remaining tasks, we then sample a balanced subset, retaining at least one and no more than two tasks per task category (e.g., image classification, training LLMs).
Consequently, we obtain a lite version of MLE-Bench conssiting of 15 tasks, which are splited into \textit{Easy}, \textit{Medium} and \textit{Hard} tiers based on human experience and the performance of prior SOTA.
We include a full list of tasks in MLE-Bench used for evaluation in Table \ref{tab:mlebench}.

\subsection{Top AI Competitions}
BELKA dataset collects from kaggle competition\footnote{\url{https://www.kaggle.com/competitions/leash-BELKA}}, which predict the the binding affinity of small molecules to specific protein targets.
It provides 2.9B molecule-protein pairs of training data.
In this paper, we sample 2.2M training rows and 590K test rows from full training data following the label distribution, and uses the AP score to evaluate the methods.

OAG dataset used in this paper is collected from KDD cup Who-is-Who incorrect assignment detection task\footnote{\url{https://www.biendata.xyz/kdd2024}}.
Given the paper assignments of each author and the metadata of each paper, this task is to detect paper assignment errors for each author. In this task, each paper contains the title, abstract, author name and corresponding organization, keywords, venue and year. The training data contains several authors and corresponding paper assignments.
For simplicity, we sample 10k papers from the given paper list, and then preserve the corresponding authors related to these papers. These authors are classified into training and test data following the label distribution used in training data.  The eveluation metric is $\sum_{1}^M w_i\times AUC_i$ where $M$ is the test author number, 
$$
w_i=\frac{\#ErrorsOfThisAuthor}{\#TotalErrors}.
$$

%% file: tables/mlebench.tex
\begin{table*}[htp]
    \caption{Full list of tasks in MLE-Bench used for evaluation in our work.}
    \label{tab:mlebench}
    \resizebox{\textwidth}{!}{
    \setlength{\tabcolsep}{0.6mm}
    \centering
\begin{tabular}{cccccc}
\toprule[1.2pt]
\textbf{Tasks}                                    & \textbf{Task Type}             & \textbf{Dataset Size (GB)} & \textbf{Metric Type} & \textbf{Original Complexity} & \textbf{Split}  \\ \midrule[1pt]
aptos2019-blindness-detection           & Image Classification & 10.22             & Max          & Low                 & Easy   \\
random-acts-of-pizza                    & Text Classification  & 0.003             & Max          & Low                 & Easy   \\
spooky-author-identification            & Text Classification  & 0.0019            & Min          & Low                 & Easy   \\
google-quest-challenge                  & Training LLMs        & 0.015             & Max          & Medium              & Easy   \\
stanford-covid-vaccine                  & Tabular              & 2.68              & Min          & High                & Easy   \\
predict-volcanic-eruptions-ingv-oe      & Signal Processing    & 31.25             & Min          & High                & Easy   \\ \midrule[1pt]
lmsys-chatbot-arena                     & Text Classification  & 0.18              & Min          & Medium              & Medium \\
us-patent-phrase-to-phrase-matching     & Text Regression      & 0.00214           & Max          & Medium              & Medium \\
mlsp-2013-birds                         & Audio Classification & 0.5851            & Max          & Low                 & Medium \\
statoil-iceberg-classifier-challenge    & Image Classification & 0.3021            & Min          & Medium              & Medium \\
tensorflow-speech-recognition-challenge & Audio Classification & 3.76              & Max          & Medium              & Medium \\ \midrule[1pt]
denoising-dirty-documents               & Image to Image       & 0.06              & Min          & Low                 & Hard   \\
new-york-city-taxi-fare-prediction      & Tabular              & 5.7               & Min          & Low                 & Hard   \\
tgs-salt-identification-challenge       & Image Segmentation   & 0.5               & Max          & Medium              & Hard   \\
ventilator-pressure-prediction          & Forecasting          & 0.7               & Min          & Medium              & Hard  \\ \bottomrule[1.2pt]
\end{tabular}
    }
\end{table*}

%% file: appendix/environment.tex
\section{Runtime Environment}
\label{app:environment}

In our experiments, LLM agents are loaded into an Ubuntu 20.04 Docker container containing the dataset prepared for each task, and an Anaconda environment pre-installed with standard Python packages for machine learning (e.g., PyTorch, scikit-learn), thereby providing all requisite dependencies for code implementation and execution.
The container runs on a compute node with 48 vCPUs, 448GB RAM, 9.6TB SSD storage, and a single NVIDIA Tesla V100 32G GPU, all of which are fully accessible to the agents.

%% file: appendix/hyperparameters.tex
\section{Hyperparameters}
\label{app:hyperparameters}

We list the detailed hyperparameters for {\ours} and AIDE in Table~\ref{tab:automind_hyper} and Table~\ref{tab:aide_hyper}, respectively.

\begin{table*}[htp]
    \centering
    \caption{Hyperparameters for {\ours}. \texttt{\&TARGET\_MODEL} is the foundation model being evaluated. \texttt{agent.search.num\_drafts} is the number of initial draft nodes. \texttt{agent.search.debug\_prob} is the probalitity of whether debug a node. \texttt{agent.search.trick\_prob} is the probalitity of whether use tricks to improve a node. \texttt{agent.search.greedy\_prob} is the probability of whether select the best node.}
        \begin{tabular}{ll}
        \toprule
            \textbf{Hyperparameter} & \textbf{Value} \\
            \midrule
            \texttt{agent.retriever.model} & \texttt{gpt-4.1-mini-2025-04-14} \\
            \texttt{agent.analyzer.model} & \texttt{gpt-4.1-mini-2025-04-14} \\
            \texttt{agent.planner.model} & \texttt{\&TARGET\_MODEL} \\
            \texttt{agent.coder.model} & \texttt{\&TARGET\_MODEL} \\
            \texttt{agent.improver.model} & \texttt{\&TARGET\_MODEL} \\
            \texttt{agent.verifier.model} & \texttt{gpt-4.1-mini-2025-04-14} \\
            \texttt{agent.steps} & \texttt{2000} \\
            \texttt{agent.search.num\_drafts} & \texttt{5} \\
            \texttt{agent.search.max\_debug\_depth} & \texttt{5} \\
            \texttt{agent.search.debug\_prob} & \texttt{1} \\
            \texttt{agent.search.trick\_prob} & \texttt{0.8} \\
            \texttt{agent.search.greedy\_prob} & \texttt{0.8} \\
            \texttt{agent.time\_limit} & \texttt{86400} \\
            \texttt{exec.timeout} & \texttt{32400} \\
        \bottomrule
        \end{tabular}
    \label{tab:automind_hyper}
\end{table*}

\begin{table*}[htp]
    \centering
    \caption{Hyperparameters for AIDE.
    \texttt{\&TARGET\_MODEL} is the foundation model being evaluated. \texttt{agent.search.num\_drafts} is the number of initial draft nodes. \texttt{agent.search.debug\_prob} is the probalitity of whether debug a node.}
        \begin{tabular}{ll}
        \toprule
            \textbf{Hyperparameter} & \textbf{Value} \\
            \midrule
            \texttt{agent.code.model} & \texttt{\&TARGET\_MODEL} \\
            \texttt{agent.feedback.model} & \texttt{gpt-4.1-mini-2025-04-14} \\
            \texttt{agent.steps} & \texttt{2000} \\
            \texttt{agent.search.num\_drafts} & \texttt{5} \\
            \texttt{agent.search.max\_debug\_depth} & \texttt{20} \\
            \texttt{agent.search.debug\_prob} & \texttt{1} \\
            \texttt{agent.time\_limit} & \texttt{86400} \\
            \texttt{exec.timeout} & \texttt{32400} \\
        \bottomrule
        \end{tabular}
    \label{tab:aide_hyper}
\end{table*}

%% file: appendix/prompts.tex
\onecolumn
\section{Prompts}
\label{app:prompts}

In this section, we showcase some of the prompts used in the full pipeline of {\ours}, which serve as a reference.

\begin{tcolorbox}[
    enhanced,
    title=Prompt for plan generation of drafting,
    breakable,
    before skip=6pt,
    after skip=6pt
]
\# Introduction

You are an expert machine learning engineer attempting a task. In order to complete this task, you need to come up with an excellent and creative plan for a solution, which will be implemented by another engineer. We will now provide a description of the task.

\# Task description

\{task\_description\}

\# Memory

Take the Memory section into consideration when proposing the solution plan, don{\textquotesingle}t propose the similar solution but keep the evaluation metric exactlty the same.

\{memory\}

\# Knowledge

Some of the tricks that have proved useful for the same type of task are provided as follows: 

\{tricks\}

You should carefully consider these tricks when designing your solution.

\# Data Analysis

\{data\_analysis\}

\# Instructions

\#\# Response format

Your response should be a detailed outline/sketch of your proposed solution in natural language. You do not need to implement the solution but you should provide enough detail for another engineer to implement it in Python code. There should be no additional headings or text in your response. Just natural language text followed by a newline.

\#\# Plan guideline

- DO NOT CHEAT ON EVALUATION. The solution should calculate the evaluation metric described in the task description on a hold-out validation set.

- If the evaluation metric is not provided, you should propose a reasonable evaluation metric for the task and calculate it.

- The solution should print the evaluation metric computed on the hold-out validation set at the last step of the solution.

- Try to come up with more modern and powerful methods to feature engineering and modelling and avoid using outdated methods. For example, if the task is a classification task, you should use modern transformer-based models instead of traditional models like CNN or LSTM.

- The solution should adopt appropriate methods to prevent model overfitting, such as data augmentation, early stopping, regularization, dropout, and others.

- Don{\textquotesingle}t suggest to do model ensembling.

- Don{\textquotesingle}t suggest to do Exploratory Data Analysis.

- Don{\textquotesingle}t suggest to do hyperparameter tuning.

- The data is already prepared and available in the {\textquotesingle}./input{\textquotesingle} directory. There is no need to unzip any files.

- The solution should use os.walk to get the paths of all available files in the {\textquotesingle}. /input{\textquotesingle} directory for data loading.

- If a {\textquotesingle}sample\_submission.csv{\textquotesingle} file existes, directly load it and use it as a template for the {\textquotesingle}submission.csv{\textquotesingle} file. The solution should save predictions on the provide unlabeled test data in the {\textquotesingle}submission.csv{\textquotesingle} file in the ./submission/ directory.

- Prefer and explicitly use GPU (CUDA) acceleration when available: move models/tensors to GPU and handle CPU fallback if CUDA is not present.

\end{tcolorbox}

\begin{tcolorbox}[
    enhanced,
    title=Prompt for plan generation of debugging,
    breakable
]
\# Introduction

You are an expert machine learning engineer attempting a task. You are provided with the plan, code and execution output of a previous solution below that had a bug and/or did not produce a submission.csv, and should improve it in order to fix the bug. For this you should first propose an reasonanle improvement and accordingly outline a detailed improved plan in natural language, which will be implemented by another engineer. We will now provide a description of the task.

\# Task description

\{task\_description\}

\# Previous Solution

\#\# Previous Plan

\{prev\_plan\}

\#\# Previous Code

\{prev\_code\}

\#\# Previous Execution Output

\{prev\_output\}

\# Data Analysis

\{data\_analysis\}

\# Instructions

\#\# Response format

First, provide a brief explanation of your reasoning for the proposed improvement to the previous plan (wrapped in <think></think>). Then, provide a detailed outline/sketch of your improved solution in natutal language based on the previous plan and your proposed improvement (wrapped in <plan></plan>). You do not need to implement the solution but you should provide enough detail for another engineer to implement it in Python code.

\#\# Installed Packages

Your solution can use any relevant machine learning packages such as: {\textquotesingle}torch-geometric==2.6.1{\textquotesingle}, {\textquotesingle}xgboost==2.1.3{\textquotesingle}, {\textquotesingle}torchvision==0.17.0{\textquotesingle}, {\textquotesingle}lightgbm==4.5.0{\textquotesingle}, {\textquotesingle}transformers==4.44.2{\textquotesingle}, {\textquotesingle}matplotlib==3.9.2{\textquotesingle}, {\textquotesingle}scipy==1.11.4{\textquotesingle}, {\textquotesingle}statsmodels==0.14.4{\textquotesingle}, {\textquotesingle}pandas==2.1.4{\textquotesingle}, {\textquotesingle}torch==2.2.0{\textquotesingle}, {\textquotesingle}optuna==4.0.0{\textquotesingle}, {\textquotesingle}timm==0.9.7{\textquotesingle}, {\textquotesingle}scikit-learn==1.2.2{\textquotesingle}, {\textquotesingle}numpy==1.26.2{\textquotesingle}, {\textquotesingle}bayesian-optimization==1.5.1{\textquotesingle}, {\textquotesingle}seaborn==0.13.2{\textquotesingle}. Feel free to use any other packages too (all packages are already installed!). For neural networks please use PyTorch because of the unavailability of TensorFlow in the environment.

\#\# Improve guideline

- You should pay attention to the execution output of the previous solution, and propose an improvement that will fix the bug.

- The improved plan should be derived by adapting the previous plan only based on the proposed improvement, while retaining other details of the previous plan.Don{\textquotesingle}t suggest to do Exploratory Data Analysis.

- Don{\textquotesingle}t suggest to do hyperparameter optimization, you should use the best hyperparameters from the previous solution.

- If a {\textquotesingle}sample\_submission.csv{\textquotesingle} file existes, directly load it and use it as a template for the {\textquotesingle}submission.csv{\textquotesingle} file. The solution should save predictions on the provide unlabeled test data in the {\textquotesingle}submission.csv{\textquotesingle} file in the ./submission/ directory.

- When describing your improved plan, do not use phrases like {\textquotesingle}the same as before{\textquotesingle} or {\textquotesingle}as in the previous plan{\textquotesingle}. Instead, fully restate all details from the previous plan that you want to retain, as subsequent implementation will not have access to the previous plan.

\end{tcolorbox}

\begin{tcolorbox}[
    enhanced,
    title=Prompt for plan generation of improving with tricks,
    breakable
]
\# Introduction

You are an expert machine learning engineer attempting a task. You are provided with the plan, code and execution output of a previous solution below and should improve it in order to further increase the test time performance. For this you should integrate integrate several useful tricks provided and accordingly outline a detailed improved plan in natural language, which will be implemented by another engineer. We will now provide a description of the task.

\# Task description

\{task\_description\}

\# Memory

Take the Memory section into consideration when proposing the solution plan, don{\textquotesingle}t propose the similar solution but keep the evaluation metric exactlty the same.

\{memory\}

\# Previous Solution

\#\# Previous Plan

\{prev\_plan\}

\#\# Previous Code

\{prev\_code\}

\#\# Previous Execution Output

\{prev\_output\}

\# Knowledge

Here are some tricks that have proved useful for the task: 

\{tricks\}

You should carefully consider these tricks when designing your solution.

\# Data Analysis

\{data\_analysis\}

\# Instructions

\#\# Response format

First, provide a brief explanation of your reasoning for the proposed improvement to the previous plan (wrapped in <think></think>). Then, provide a detailed outline/sketch of your improved solution in natutal language based on the previous plan and your proposed improvement (wrapped in <plan></plan>). You do not need to implement the solution but you should provide enough detail for another engineer to implement it in Python code.

\#\# Installed Packages

Your solution can use any relevant machine learning packages such as: {\textquotesingle}torch-geometric==2.6.1{\textquotesingle}, {\textquotesingle}xgboost==2.1.3{\textquotesingle}, {\textquotesingle}torchvision==0.17.0{\textquotesingle}, {\textquotesingle}lightgbm==4.5.0{\textquotesingle}, {\textquotesingle}transformers==4.44.2{\textquotesingle}, {\textquotesingle}matplotlib==3.9.2{\textquotesingle}, {\textquotesingle}scipy==1.11.4{\textquotesingle}, {\textquotesingle}statsmodels==0.14.4{\textquotesingle}, {\textquotesingle}pandas==2.1.4{\textquotesingle}, {\textquotesingle}torch==2.2.0{\textquotesingle}, {\textquotesingle}optuna==4.0.0{\textquotesingle}, {\textquotesingle}timm==0.9.7{\textquotesingle}, {\textquotesingle}scikit-learn==1.2.2{\textquotesingle}, {\textquotesingle}numpy==1.26.2{\textquotesingle}, {\textquotesingle}bayesian-optimization==1.5.1{\textquotesingle}, {\textquotesingle}seaborn==0.13.2{\textquotesingle}. Feel free to use any other packages too (all packages are already installed!). For neural networks please use PyTorch because of the unavailability of TensorFlow in the environment.

\#\# Improve guideline

- You should focus ONLY on integrating the provided tricks in the knowledge section into the previous solution to fully leverage their potentials.

- Make sure to fully integrate these tricks into your plan while preserving as much details as possible.

- Ensure that your plan clearly demonstrates the functions and specifics of the tricks.

- Identify the key areas in the previous solution where the knowledge can be applied.

- Suggest specific changes or additions to the code or plan based on the knowledge provided, and avoid unnecessary modifications irrelevant to the tricks.

- If a {\textquotesingle}sample\_submission.csv{\textquotesingle} file existes, directly load it and use it as a template for the {\textquotesingle}submission.csv{\textquotesingle} file. The solution should save predictions on the provide unlabeled test data in the {\textquotesingle}submission.csv{\textquotesingle} file in the ./submission/ directory.

- When describing your improved plan, do not use phrases like {\textquotesingle}the same as before{\textquotesingle} or {\textquotesingle}as in the previous plan{\textquotesingle}. Instead, fully restate all details from the previous plan that you want to retain, as subsequent implementation will not have access to the previous plan.

\end{tcolorbox}

\begin{tcolorbox}[
    enhanced,
    title=Prompt for plan generation of improving without tricks,
    breakable
]
\# Introduction

You are an expert machine learning engineer attempting a task. You are provided with the plan, code and execution output of a previous solution below and should improve it in order to further increase the test time performance. For this you should first propose a reasonable improvement and accordingly outline a detailed improved plan in natural language, which will be implemented by another engineer. We will now provide a description of the task.

\# Task description

\{task\_description\}

\# Memory

Take the Memory section into consideration when proposing the solution plan, don{\textquotesingle}t propose the similar solution but keep the evaluation metric exactlty the same.

\{memory\}

\# Previous Solution

\#\# Previous Plan

\{prev\_plan\}

\#\# Previous Code

\{prev\_code\}

\#\# Previous Execution Output

\{prev\_output\}

\# Data Analysis

\{data\_analysis\}

\# Instructions

\#\# Response format

First, provide a brief explanation of your reasoning for the proposed improvement to the previous plan (wrapped in <think></think>). Then, provide a detailed outline/sketch of your improved solution in natutal language based on the previous plan and your proposed improvement (wrapped in <plan></plan>). You do not need to implement the solution but you should provide enough detail for another engineer to implement it in Python code.

\#\# Installed Packages

Your solution can use any relevant machine learning packages such as: {\textquotesingle}torch-geometric==2.6.1{\textquotesingle}, {\textquotesingle}xgboost==2.1.3{\textquotesingle}, {\textquotesingle}torchvision==0.17.0{\textquotesingle}, {\textquotesingle}lightgbm==4.5.0{\textquotesingle}, {\textquotesingle}transformers==4.44.2{\textquotesingle}, {\textquotesingle}matplotlib==3.9.2{\textquotesingle}, {\textquotesingle}scipy==1.11.4{\textquotesingle}, {\textquotesingle}statsmodels==0.14.4{\textquotesingle}, {\textquotesingle}pandas==2.1.4{\textquotesingle}, {\textquotesingle}torch==2.2.0{\textquotesingle}, {\textquotesingle}optuna==4.0.0{\textquotesingle}, {\textquotesingle}timm==0.9.7{\textquotesingle}, {\textquotesingle}scikit-learn==1.2.2{\textquotesingle}, {\textquotesingle}numpy==1.26.2{\textquotesingle}, {\textquotesingle}bayesian-optimization==1.5.1{\textquotesingle}, {\textquotesingle}seaborn==0.13.2{\textquotesingle}. Feel free to use any other packages too (all packages are already installed!). For neural networks please use PyTorch because of the unavailability of TensorFlow in the environment.

\#\# Improve guideline

- You should conduct only one expert-level actionable improvement to the previous solution.

- This improvement should be atomic so that the effect of the improved solution can be experimentally evaluated.

- The improved plan should be derived by adapting the previous plan only based on the proposed improvement, while retaining other details of the previous plan.

- Don't suggest to do Exploratory Data Analysis.

- Don't suggest to do hyperparameter optimization, you should use the best hyperparameters from the previous solution.

- If a {\textquotesingle}sample\_submission.csv{\textquotesingle} file existes, directly load it and use it as a template for the {\textquotesingle}submission.csv{\textquotesingle} file. The solution should save predictions on the provide unlabeled test data in the {\textquotesingle}submission.csv{\textquotesingle} file in the ./submission/ directory.

- When describing your improved plan, do not use phrases like {\textquotesingle}the same as before{\textquotesingle} or {\textquotesingle}as in the previous plan{\textquotesingle}. Instead, fully restate all details from the previous plan that you want to retain, as subsequent implementation will not have access to the previous plan.

\end{tcolorbox}

\begin{tcolorbox}[
    enhanced,
    title=Prompt for complexity scorer,
    breakable
]

\# Introduction

You are an expert machine learning engineer attempting a task. In order to complete this task, you are given a discription of the task and a solution plan proposed by another engineer and need to assess the complexity of the task and the proposed solution. We will now provide a description of the task.

\# Task description

\{task\_description\}

\# Proposed Solution

\{proposed\_solution\}

\# Data Analysis

\{data\_analysis\}

\# Instructions

\#\# Response format

First, provide a brief explanation of your reasoning for the assessment of the complexity of the task and the proposed solution (wrapped in <think></think>). Then, provide ONLY ONE average complexity score as floating point number between 1 and 5, which can contain 0.5 points (wrapped in <score></score>).

\#\# Task complexity scoring criteria

- 5 = Extremely complex and cutting-edge task with high levels of innovation required. This involves solving a unique or highly specialized problem that may push the boundaries of existing knowledge or technology.

- 4 = Complex task that involves advanced techniques or methodologies, requiring considerable expertise in the domain, such as building novel algorithms, optimization methods, or handling advanced data.

- 3 = Moderately complex task that requires significant problem-solving, such as integrating different methods or creating custom algorithms for specific use cases.

- 2 = Simple task with some level of complexity, such as basic algorithms that need some degree of fine-tuning or adjustment to meet the specific requirements of the project.

- 1 = Very simple task that requires minimal effort, such as basic data manipulation or applying standard algorithms without any customization.

\#\# Proposed solution complexity scoring criteria

- 5 = A groundbreaking or transformative solution that pushes the envelope in the field. It introduces a novel approach that is scalable, efficient, and offers long-term value or sets a new standard.

- 4 = A highly original and effective solution that shows a deep understanding of the problem domain and offers a significant contribution to the field. The solution is well-optimized and efficient.

- 3 = An original and creative solution with a reasonable level of complexity. It involves designing and implementing custom solutions or combining existing methods in a new way.

- 2 = A somewhat original solution that involves adapting existing tools or methods with some customization to meet the needs of the project. There may be room for optimization or improvement.

- 1 = Very basic solution, perhaps using standard, off-the-shelf tools with minimal adaptation, lacking originality or novel contributions.

\#\# Complexity scoring guideline

- Evaluate the complexity of the task and the proposed solution, and assign a score between 1 and 5.

- Assign an average score between 1 and 5, consider factors such as the task{\textquotesingle}s complexity, the proposed solution, the dataset size, and the time and hardware resources required for implementation and execution.

\end{tcolorbox}

\begin{tcolorbox}[
    enhanced,
    title=Prompt for code implementation through one-pass coding,
    breakable
]

\# Introduction

You are an expert machine learning engineer attempting a task. In order to complete this task, you are given a discription of the task and a solution plan proposed by another engineer and need to assess the complexity of the task and the proposed solution. We will now provide a description of the task.

\# Task description

\{task\_description\}

\# Proposed Solution

\{proposed\_solution\}

\# Data Analysis

\{data\_analysis\}

\# Instructions

\#\# Response format

Your response should be a single markdown code block (wrapped in {\textquotesingle}{\textquotesingle}{\textquotesingle}) which implements this solution plan and prints out and save the evaluation metric.

\#\# Installed Packages

Your solution can use any relevant machine learning packages such as: {\textquotesingle}torch-geometric==2.6.1{\textquotesingle}, {\textquotesingle}xgboost==2.1.3{\textquotesingle}, {\textquotesingle}torchvision==0.17.0{\textquotesingle}, {\textquotesingle}lightgbm==4.5.0{\textquotesingle}, {\textquotesingle}transformers==4.44.2{\textquotesingle}, {\textquotesingle}matplotlib==3.9.2{\textquotesingle}, {\textquotesingle}scipy==1.11.4{\textquotesingle}, {\textquotesingle}statsmodels==0.14.4{\textquotesingle}, {\textquotesingle}pandas==2.1.4{\textquotesingle}, {\textquotesingle}torch==2.2.0{\textquotesingle}, {\textquotesingle}optuna==4.0.0{\textquotesingle}, {\textquotesingle}timm==0.9.7{\textquotesingle}, {\textquotesingle}scikit-learn==1.2.2{\textquotesingle}, {\textquotesingle}numpy==1.26.2{\textquotesingle}, {\textquotesingle}bayesian-optimization==1.5.1{\textquotesingle}, {\textquotesingle}seaborn==0.13.2{\textquotesingle}. Feel free to use any other packages too (all packages are already installed!). For neural networks please use PyTorch because of the unavailability of TensorFlow in the environment.

\#\# Code guideline

- The code should **implement the proposed solution** and **print the value of the evaluation metric computed on a hold-out validation set**,

- **AND MOST IMPORTANTLY SAVE PREDICTIONS ON THE PROVIDED UNLABELED TEST DATA IN A {\textquotesingle}submission.csv{\textquotesingle} FILE IN THE ./submission/ DIRECTORY.**

- The code should save the evaluation metric computed on the hold-out validation set in a {\textquotesingle}eval\_metric.txt{\textquotesingle} file in the ./submission/ directory.

- The code should be a single-file python program that is self-contained and can be executed as-is.

- No parts of the code should be skipped, don{\textquotesingle}t terminate the code before finishing the script.

- DO NOT WRAP THE CODE IN A MAIN FUNCTION, BUT WRAP ALL CODE in the {\textquotesingle}\_\_main\_\_{\textquotesingle} module, or it cannot be executed successfully.

- All class initializations and computational routines MUST BE WRAPPED in {\textquotesingle}if \_\_name\_\_ == "\_\_main\_\_":{\textquotesingle}.

- DO NOT USE MULTIPROCESSING OR SET {\textquotesingle}num\_workers{\textquotesingle} IN DATA LOADER, as it may cause the container to crash.

- Your response should only contain a single code block.

- All input data is already prepared and available in the {\textquotesingle}./input{\textquotesingle} directory. There is no need to unzip any files.

- DO NOT load data from "./data" directory, it is not available in the environment.

- Do not save any intermediate or temporary files through {\textquotesingle}torch.save{\textquotesingle} or {\textquotesingle}pickle.dump{\textquotesingle}.

- Try to accelerate the model training process if any GPU is available.

- DO NOT display progress bars. If you have to use function intergrated with progress bars, disable progress bars or using the appropriate parameter to silence them.

- Don{\textquotesingle}t do Exploratory Data Analysis.

- Avoid printing detailed model architecture information unless debugging. When debugging model issues, use concise shape tracking during forward pass to quickly identify problematic layers without verbose model summaries.

- When debugging data-related errors, please refer to the data analysis section first for insights about data structure and format.

- **DO NOT HARDCODE OR FAKE THE EVALUATION METRIC VALUE. The metric must be computed from actual model performance on validation data.**

\end{tcolorbox}

\begin{tcolorbox}[
    enhanced,
    title=Prompt for stepwise decomposition,
    breakable
]

\# Introduction

You are an expert machine learning engineer attempting a task. In order to complete this task, you are given the proposed solution and supposed to decompose it into multiple steps. We will now provide a description of the task.

\# Task description

\{task\_description\}

\# Proposed Solution

\{proposed\_solution\}

\# Instructions

\#\# Response format

- Your response should be a single JSON code block (wrapped in {\textquotesingle}{\textquotesingle}{\textquotesingle}) which contains the decomposition steps of the proposed solution.

- The generated JSON should have the following format: 

\{

\quad"decomposed steps": [

\quad\quad\{

\quad\quad\quad"step": "Name of the step",

\quad\quad\quad"details": "Detailed description of the step",

\quad\quad\},

\quad\quad...

\quad],

\}

\#\# Solution decomposition guideline

- You should decompose the proposed solution into multiple steps, and provide detailed descriptions of each step.

- DO NOT MODIFY THE PROPOSED SOLUTION. In the description of each step, you should keep as many details of the proposed solution as possible, especially the exact hyperparameters and sample code.

- DO NOT CHEAT ON EVALUATION. The solution should calculate the evaluation metric described in the task description on a hold-out validation set.

- If the evaluation metric is not provided, you should propose a reasonable evaluation metric for the task and calculate it.

- The solution should save the evaluation metric computed on the hold-out validation set in a {\textquotesingle}eval\_metric.txt{\textquotesingle} file in the ./submission/ directory.

- The solution should use os.walk to get the paths of all available files in the {\textquotesingle}. /input{\textquotesingle} directory for data loading.

- If a sample\_submission.csv file existes, directly load it and use it as a template for the {\textquotesingle}submission.csv{\textquotesingle} file. The solution should save predictions on the provide unlabeled test data in the {\textquotesingle}submission.csv{\textquotesingle} file in the ./submission/ directory.

- You should **print the value of the evaluation metric computed on a hold-out validation set** in the last step of the decomposed steps.

- Don{\textquotesingle}t do Exploratory Data Analysis in the decomposition steps.

- If you find improvements suggestions in the plan, you should take them in serious consideration and include them in the decomposition steps.

- You do not need to implement the code in the decomposed steps.

- Note that the order of the decomposed steps determines the order in which the code is implemented and executed.

\end{tcolorbox}

\begin{tcolorbox}[
    enhanced,
    title=Prompt for code implementation through stepwise coding,
    breakable
]

\# Introduction

You are an expert machine learning engineer attempting a task. In order to complete this task, you are given the code for previous steps and need to implement the current step of a natural language solution plan proposed by another engineer in Python code. We will now provide a description of the task.

\# Task description

\{task\_description\}

\# Current Step

\{current\_step\}

\# Previous Steps Code

You should continue the following code for previous steps to implement the current step of the solution plan, but do not repeat it: 

\{prev\_steps\}

\# Data Analysis

\{data\_analysis\}

\# Instructions

\#\# Response format

First, provide suggestions for the current step based on the previous steps and the failed last try step if provided (wrapped in <think></think>). Then, provide a single markdown code block (wrapped in {\textquotesingle}{\textquotesingle}{\textquotesingle}) which implements the current step of a solution plan.

\#\# Installed Packages

Your solution can use any relevant machine learning packages such as: {\textquotesingle}torch-geometric==2.6.1{\textquotesingle}, {\textquotesingle}xgboost==2.1.3{\textquotesingle}, {\textquotesingle}torchvision==0.17.0{\textquotesingle}, {\textquotesingle}lightgbm==4.5.0{\textquotesingle}, {\textquotesingle}transformers==4.44.2{\textquotesingle}, {\textquotesingle}matplotlib==3.9.2{\textquotesingle}, {\textquotesingle}scipy==1.11.4{\textquotesingle}, {\textquotesingle}statsmodels==0.14.4{\textquotesingle}, {\textquotesingle}pandas==2.1.4{\textquotesingle}, {\textquotesingle}torch==2.2.0{\textquotesingle}, {\textquotesingle}optuna==4.0.0{\textquotesingle}, {\textquotesingle}timm==0.9.7{\textquotesingle}, {\textquotesingle}scikit-learn==1.2.2{\textquotesingle}, {\textquotesingle}numpy==1.26.2{\textquotesingle}, {\textquotesingle}bayesian-optimization==1.5.1{\textquotesingle}, {\textquotesingle}seaborn==0.13.2{\textquotesingle}. Feel free to use any other packages too (all packages are already installed!). For neural networks please use PyTorch because of the unavailability of TensorFlow in the environment.

\#\# Code guideline

- You should implement the current step of the solution plan based on the previous steps and the failed last try step if provided.

- **You should ONLY implement the code for the current step of the solution plan, rather than the entire solution plan.**

- DO NOT MODIFY THE CURRENT STEP. You should implement the current step exactly as it is.

- You should **print the value of the evaluation metric computed on a hold-out validation set** if it is calculated in the current step.

- You should save the evaluation metric computed on the hold-out validation set in an {\textquotesingle}eval\_metric.txt{\textquotesingle} file in the {\textquotesingle}./submission/{\textquotesingle} directory if it is calculated in the current step.

- DO NOT PRINT ANYTHING ELSE IN THE CODE, except for the evaluation metric and a concise completion message for the current step.

- **DO NOT REPEAT the code for previous steps; you should only import them from {\textquotesingle}prev\_steps.py{\textquotesingle}.**

- DO NOT REPETITIVELY IMPORT THE SAME MODULES ALREADY USED IN PREVIOUS STEPS, but you may import additional modules if needed.

- **AND MOST IMPORTANTLY SAVE PREDICTIONS ON THE PROVIDED UNLABELED TEST DATA IN A {\textquotesingle}submission.csv{\textquotesingle} FILE IN THE {\textquotesingle}./submission/{\textquotesingle} DIRECTORY** if predictions are involved in the current step.

- You can reference the based code to implement the current step, but do not completely repeat it.

- **DO NOT HARDCODE OR FAKE THE EVALUATION METRIC VALUE.** It must be computed from actual model performance on validation data.

- The code should be a single-file Python program that is self-contained and can be executed as-is.

- DO NOT wrap the code in a main function, BUT WRAP ALL CODE in the {\textquotesingle}\_\_main\_\_{\textquotesingle} module, or it cannot be executed successfully.

- All class initializations and computational routines MUST BE WRAPPED in {\textquotesingle}if \_\_name\_\_ == "\_\_main\_\_":{\textquotesingle}.

- DO NOT USE MULTIPROCESSING OR SET {\textquotesingle}num\_workers{\textquotesingle} in any DataLoader.

- No parts of the code should be skipped; do not terminate early.

- All input data is already prepared and available in the {\textquotesingle}./input{\textquotesingle} directory. There is no need to unzip any files.

- DO NOT load data from the {\textquotesingle}./data{\textquotesingle} directory (not available).

- Do not save any intermediate or temporary files through {\textquotesingle}torch.save{\textquotesingle} or {\textquotesingle}pickle.dump{\textquotesingle}.

- Feel free to use GPU in any stage if it is available.

- DO NOT display progress bars; disable them or silence via parameters.

- Don't do Exploratory Data Analysis.

- Avoid printing detailed model architecture information unless debugging. For debugging, use concise tensor shape tracking.

- When debugging data-related errors, first refer to the data analysis section for structure/format insights.

\end{tcolorbox}

\begin{tcolorbox}[
    enhanced,
    title=Prompt for debugging through one-pass coding,
    breakable
]

\# Introduction

You are debugging a failed ML code step. Use precise SEARCH/REPLACE format to fix errors.

\# Task description

\{task\_description\}

\# Improved Solution Plan

\{iproved\_solution\_plan\}

\# Failed Code

\{failed\_code\}

\# Instructions

- Diff Format

\begin{adjustwidth}{2em}{0pt}
Use EXACT SEARCH/REPLACE format:  \\
\texttt{\textless{}\textless{}\textless{}\textless{}\textless{}\textless{}\textless{}} SEARCH  \\
\# exact code to replace (must match exactly)  \\
=======  \\
\# new code  \\
\texttt{\textgreater{}\textgreater{}\textgreater{}\textgreater{}\textgreater{}\textgreater{}\textgreater{}} REPLACE  

The SEARCH block must match the code exactly, including whitespace.  
Focus on targeted fixes, not full rewrites.  
You can make multiple changes with multiple diff blocks.  
Explain the reasoning for each change.
\end{adjustwidth}

- The code should **implement the proposed solution** and **print the value of the evaluation metric computed on a hold-out validation set**,

- **AND MOST IMPORTANTLY SAVE PREDICTIONS ON THE PROVIDED UNLABELED TEST DATA IN A {\textquotesingle}submission.csv{\textquotesingle} FILE IN THE ./submission/ DIRECTORY.**

- The code should save the evaluation metric computed on the hold-out validation set in a {\textquotesingle}eval\_metric.txt{\textquotesingle} file in the ./submission/ directory.

- DO NOT HARDCODE OR FAKE THE EVALUATION METRIC VALUE. The metric must be computed from actual model performance on validation data.

- The code should be a single-file python program that is self-contained and can be executed as-is.

- DO NOT WRAP THE CODE IN A MAIN FUNCTION, BUT WRAP ALL CODE in the {\textquotesingle}\_\_main\_\_{\textquotesingle} module, or it cannot be executed successfully.

- All class initializations and computational routines MUST BE WRAPPED in {\textquotesingle}if \_\_name\_\_ == "\_\_main\_\_":{\textquotesingle}.

- DO NOT USE MULTIPROCESSING OR SET {\textquotesingle}num\_workers{\textquotesingle} IN DATA LOADER, as it may cause the container to crash.

- No parts of the code should be skipped, don{\textquotesingle}t terminate the code before finishing the script.

- All input data is already prepared and available in the {\textquotesingle}./input{\textquotesingle} directory. There is no need to unzip any files.

- DO NOT load data from {\textquotesingle}./data{\textquotesingle} directory, it is not available in the environment.

- Do not save any intermediate or temporary files through {\textquotesingle}torch.save{\textquotesingle} or {\textquotesingle}pickle.dump{\textquotesingle}.

- Feel free to use GPU in any stage if it is available.

- DO NOT display progress bars. If you have to use function integrated with progress bars, disable progress bars or use the appropriate parameter to silence them.

- Don{\textquotesingle}t do Exploratory Data Analysis.

- Avoid printing detailed model architecture information unless debugging. When debugging model issues, use concise shape tracking during forward pass to quickly identify problematic layers without verbose model summaries.

- When debugging data-related errors, please refer to the data analysis section first for insights about data structure and format.

\end{tcolorbox}

\begin{tcolorbox}[
    enhanced,
    title=Prompt for debugging in stepwise coding,
    breakable
]

\# Introduction

You are debugging a failed ML code step. Use precise SEARCH/REPLACE format to fix errors.

\# Current Step

\{current\_step\}

\# Failed Code

\{failed\_code\}

\# Error Output

\{error\_output\}

\# Instructions

- IMPORTANT

\begin{adjustwidth}{2em}{0pt}
- You can ONLY modify the Failed Code shown above

- Do NOT search for code from previous steps

- Your SEARCH blocks must match code in the Failed Code section exactly

- Focus only on fixing the current step's implementation

\end{adjustwidth}

- Diff Format

\begin{adjustwidth}{2em}{0pt}
Use EXACT SEARCH/REPLACE format:  \\
\texttt{\textless{}\textless{}\textless{}\textless{}\textless{}\textless{}\textless{}}
 SEARCH  \\
\# exact code to replace (must match exactly)  \\
=======  \\
\# new code  \\
\texttt{\textgreater{}\textgreater{}\textgreater{}\textgreater{}\textgreater{}\textgreater{}\textgreater{}} REPLACE 

The SEARCH block must match the code exactly, including whitespace.  
Focus on targeted fixes, not full rewrites.  
You can make multiple changes with multiple diff blocks.  
Explain the reasoning for each change.
\end{adjustwidth}

- The code should **implement the proposed solution** and **print the value of the evaluation metric computed on a hold-out validation set**,

- **AND MOST IMPORTANTLY SAVE PREDICTIONS ON THE PROVIDED UNLABELED TEST DATA IN A {\textquotesingle}submission.csv{\textquotesingle} FILE IN THE ./submission/ DIRECTORY.**

- The code should save the evaluation metric computed on the hold-out validation set in a {\textquotesingle}eval\_metric.txt{\textquotesingle} file in the ./submission/ directory.

- DO NOT HARDCODE OR FAKE THE EVALUATION METRIC VALUE. The metric must be computed from actual model performance on validation data.

- The code should be a single-file python program that is self-contained and can be executed as-is.

- DO NOT WRAP THE CODE IN A MAIN FUNCTION, BUT WRAP ALL CODE in the {\textquotesingle}\_\_main\_\_{\textquotesingle} module, or it cannot be executed successfully.

- All class initializations and computational routines MUST BE WRAPPED in {\textquotesingle}if \_\_name\_\_ == "\_\_main\_\_":{\textquotesingle}.

- DO NOT USE MULTIPROCESSING OR SET {\textquotesingle}num\_workers{\textquotesingle} IN DATA LOADER, as it may cause the container to crash.

- No parts of the code should be skipped, don{\textquotesingle}t terminate the code before finishing the script.

- All input data is already prepared and available in the {\textquotesingle}./input{\textquotesingle} directory. There is no need to unzip any files.

- DO NOT load data from {\textquotesingle}./data{\textquotesingle} directory, it is not available in the environment.

- Do not save any intermediate or temporary files through {\textquotesingle}torch.save{\textquotesingle} or {\textquotesingle}pickle.dump{\textquotesingle}.

- Feel free to use GPU in any stage if it is available.

- DO NOT display progress bars. If you have to use function integrated with progress bars, disable progress bars or use the appropriate parameter to silence them.

- Don{\textquotesingle}t do Exploratory Data Analysis.

- Avoid printing detailed model architecture information unless debugging. When debugging model issues, use concise shape tracking during forward pass to quickly identify problematic layers without verbose model summaries.

- When debugging data-related errors, please refer to the data analysis section first for insights about data structure and format.

\# Previous Steps Code

Continue from (DO NOT MODIFY): \{prev\_code\}

\end{tcolorbox}

\begin{tcolorbox}[
    enhanced,
    title=Prompt for output veirification,
    breakable
]

\# Introduction

You are an expert machine learning engineer attempting a task. You have written code to solve this task and now need to evaluate the output of the code execution. You should determine if there were any bugs as well as report the empirical findings.

\# Task description

\{task\_description\}

\# Code

\{code\}

\# Execution Output

\{execution\_output\}

\# Tool

\{

    \quad "type": "function",
    
    \quad"function": \{
    
        \quad\quad "name": "submission\_verify",
        
        \quad\quad"description": "Verify the execution output of the written code.",
        
        \quad\quad"parameters": \{
        
            \quad\quad\quad"type": "object",
                
            \quad\quad\quad"properties": \{
                
                \quad\quad\quad\quad"is\_bug": \{
                
                    \quad\quad\quad\quad\quad"type": "boolean",
                    
                    \quad\quad\quad\quad\quad"description": "true if the output log shows that the execution failed or has some bug, otherwise false.",
                    
                \quad\quad\quad\quad\},
                \quad\quad\quad\quad"is\_overfitting": \{
                
                    \quad\quad\quad\quad\quad"type": "boolean",
                    
                    \quad\quad\quad\quad\quad"description": "true if the output log shows that the model is overfitting or validation metric is much worse than the training metric or validation loss is increasing, otherwise false. ",
                \quad\quad\quad\quad\},
                \quad\quad\quad\quad"has\_csv\_submission": \{
                
                    \quad\quad\quad\quad\quad"type": "boolean",
                    
                    \quad\quad\quad\quad\quad"description": "true if the code saves the predictions on the test data in a {\textquotesingle}submission.csv{\textquotesingle} file in the {\textquotesingle}./submission/{\textquotesingle} directory, otherwise false. Note that the file MUST be saved in the ./submission/ directory for this to be evaluated as true, otherwise it should be evaluated as false. You can assume the ./submission/ directory exists and is writable.",
                    
                \quad\quad\quad\quad\},
                
                \quad\quad\quad\quad"summary": \{
                
                    \quad\quad\quad\quad\quad"type": "string",
                    
                    \quad\quad\quad\quad\quad"description": "write a short summary (2-3 sentences) describing the empirical findings. Alternatively mention if there is a bug or the submission.csv was not properly produced. You do not need to suggest fixes or improvements.",
                    
                \quad\quad\quad\quad\},
                
                \quad\quad\quad\quad"metric": \{
                
                    \quad\quad\quad\quad\quad"type": "number",
                    
                    \quad\quad\quad\quad\quad"description": "If the code ran successfully, report the value of the validation metric. Otherwise, leave it null.",
                    
                \quad\quad\quad\quad\},
                
                \quad\quad\quad\quad"lower\_is\_better": \{
                
                    \quad\quad\quad\quad\quad"type": "boolean",
                    
                    \quad\quad\quad\quad\quad"description": "true if the metric should be minimized (i.e. a lower metric value is better, such as with MSE), false if the metric should be maximized (i.e. a higher metric value is better, such as with accuracy).",
                    
                \quad\quad\quad\quad\},
                
            \quad\quad\quad\},
            
            \quad\quad\quad"required": ["is\_bug", "is\_overfitting", "has\_csv\_submission", "summary", "metric", "lower\_is\_better"],
            
        \quad\quad\},
        
    \quad\},

\}

\end{tcolorbox}

\twocolumn